%% file: main.tex
\documentclass[runningheads]{llncs}

\usepackage{eccv}

\usepackage{eccvabbrv}

\usepackage{graphicx}
\usepackage{booktabs}
\usepackage{multirow}

\usepackage[accsupp]{axessibility}  

\usepackage{nicefrac}
\usepackage{float}
\usepackage{wrapfig}
\usepackage{authoraftertitle}

\usepackage{array}
\newcolumntype{H}{>{\setbox0=\hbox\bgroup}c<{\egroup}@{}}

\usepackage{pifont}

\usepackage{hyperref}

\usepackage{orcidlink}

\begin{document}

\title{Top-GAP: Integrating Size Priors in CNNs for more Interpretability, Robustness, and Bias Mitigation}

\titlerunning{Top-GAP}

\author{Lars Nieradzik\inst{1}\orcidlink{0000-0002-7523-5694} \and
Henrike Stephani\inst{1}\orcidlink{0000-0002-9821-1636} \and
Janis Keuper\inst{2}\orcidlink{0000-0002-1327-1243}}

\authorrunning{Nieradzik et al.}

\institute{Image Processing, Fraunhofer ITWM, Fraunhofer Platz 1, Kaiserslautern, 67663, Germany\\\email{\{lars.nieradzik,henrike.stephani\}@itwm.fraunhofer.de} \and Institute for Machine Learning and Analysis, Offenburg University, Badstr. 24, Offenburg, 77652, Germany\\\email{keuper@imla.ai}}

\maketitle

\begin{abstract}
This paper introduces Top-GAP, a novel regularization technique that enhances the explainability and robustness of convolutional neural networks. By constraining the spatial size of the learned feature representation, our method forces the network to focus on the most salient image regions, effectively reducing background influence.
Using adversarial attacks and the Effective Receptive Field, we show that Top-GAP directs more attention towards object pixels rather than the background. This leads to enhanced interpretability and robustness. We achieve over 50\% robust accuracy on CIFAR-10 with PGD $\epsilon=\nicefrac{8}{255}$ and $20$ iterations while maintaining the original clean accuracy. Furthermore, we see increases of up to 5\% accuracy against distribution shifts. Our approach also yields more precise object localization, as evidenced by up to 25\% improvement in Intersection over Union (IOU) compared to methods like GradCAM and Recipro-CAM.
  \keywords{Class activation maps \and Robustness \and Adversarial attacks}
\end{abstract}

\section{Introduction}

Modern computer vision has made remarkable progress with the proliferation of Deep Learning, particularly convolutional neural networks (CNNs). These networks have demonstrated unprecedented capabilities in tasks ranging from image classification to semantic segmentation \cite{zarandy2015overview}. However, the explainability of these models remains a critical problem.

\begin{figure}
    \centering
    \begin{subfigure}{0.4\textwidth}
        \centering
        \includegraphics[width=\textwidth]{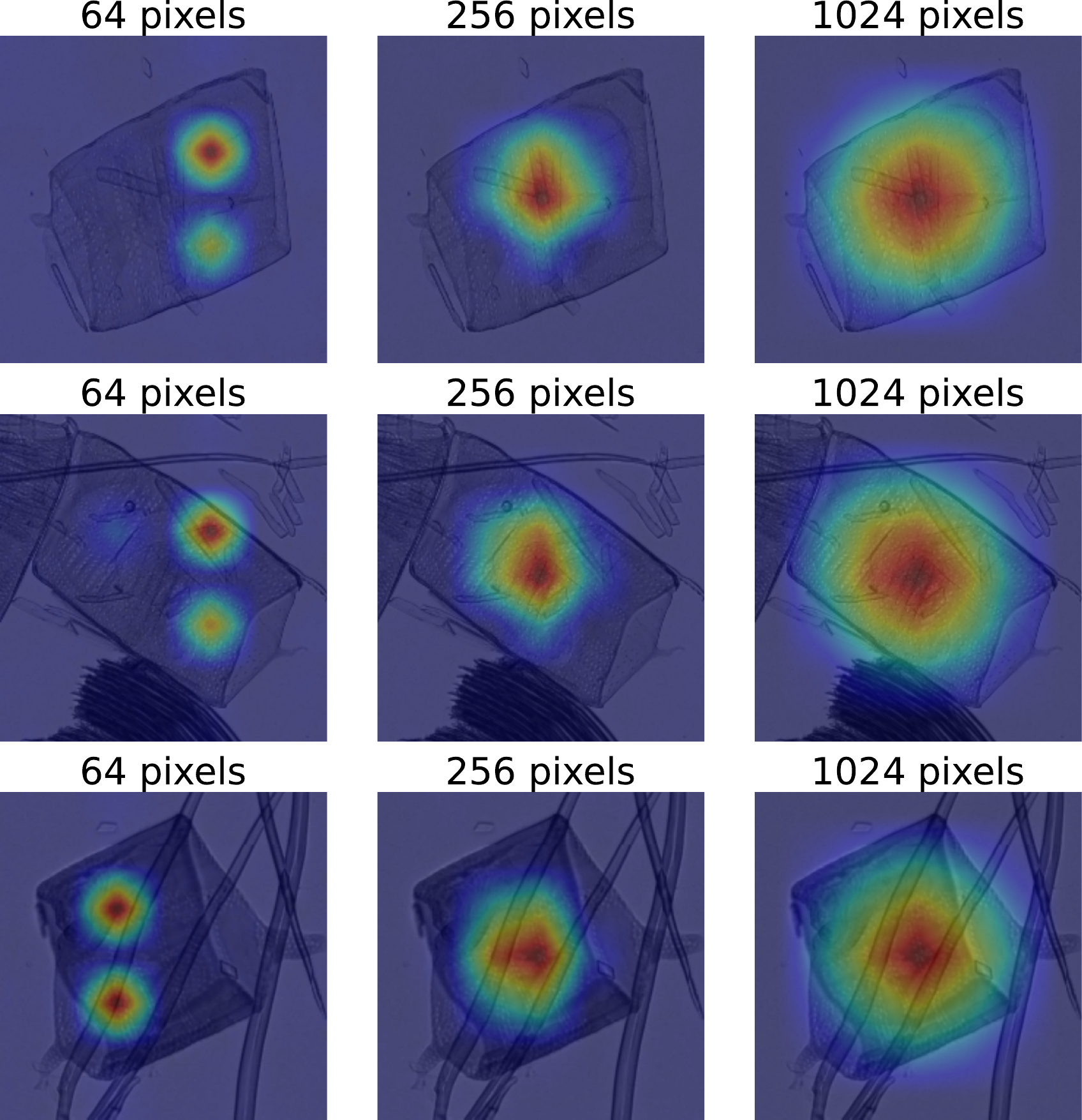}
        \caption{Wood identification \cite{nieradzik2023automating}}
        \label{fig:subfiga}
    \end{subfigure}
    \hspace{1cm}
    \begin{subfigure}{0.4\textwidth}
        \centering
        \includegraphics[width=\textwidth]{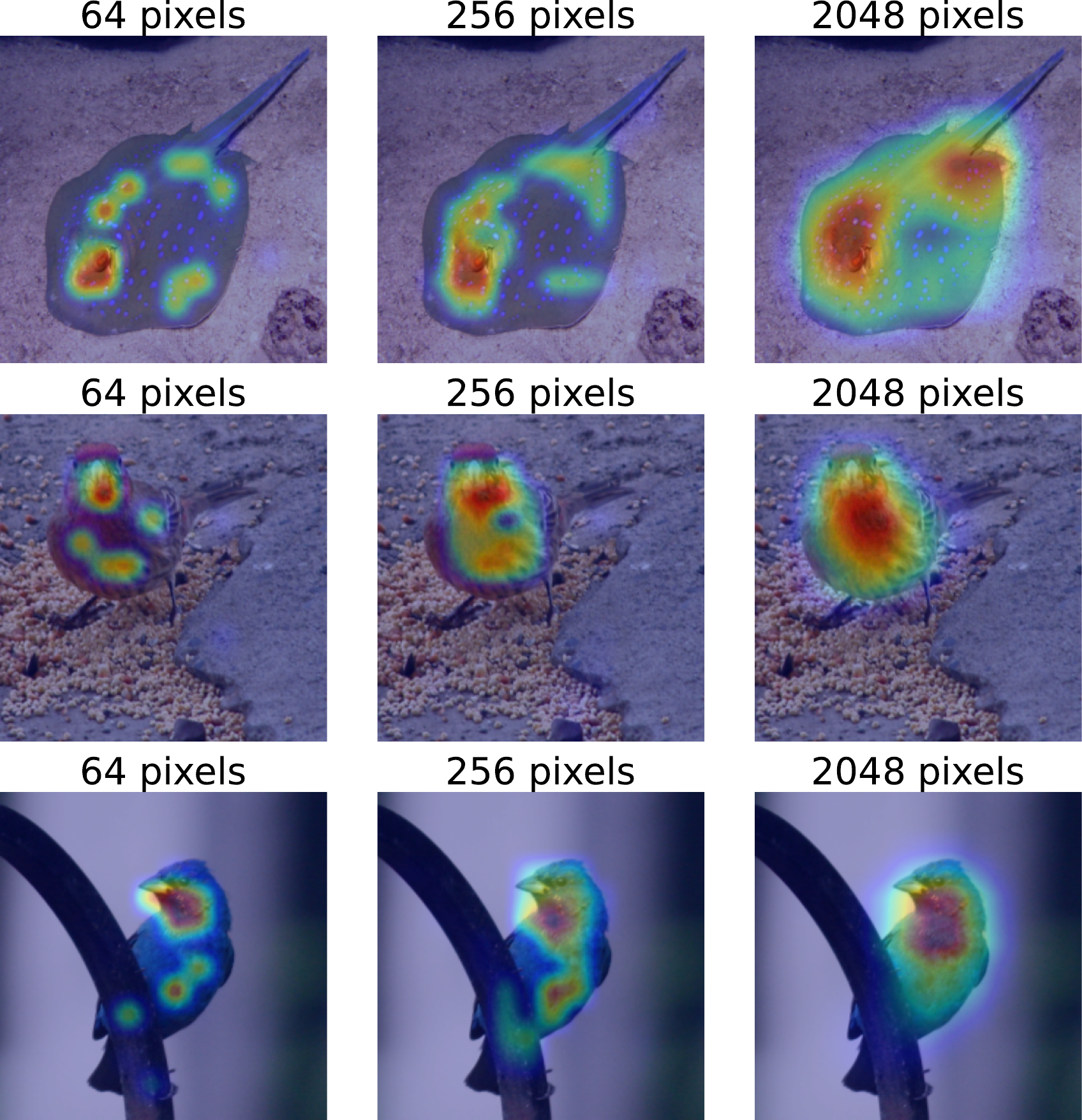}
        \caption{ImageNet \cite{5206848}}
        \label{fig:subfigb}
    \end{subfigure}
    \caption{Example images from a biological classification dataset (a) and ImageNet (b), where we limit the locations in the output feature map that the CNN can use to make predictions. Increasing the allowed pixel count leads to more pixels being highlighted in the class activation map (CAM). If the object size is not known or variable, the pixel constraint with the highest accuracy can be selected.}
    \label{fig:mainfigure}
\end{figure}

Many previous attempts to improve explainability have focused on improving class activation maps of the already trained networks. We propose a different approach that focuses on a novel method to regularize the network during training. A constraint is added to the training procedure that limits the spatial size of the learned feature representation which a neural network can use for a prediction. Unlike \cite{pathak2015constrained}, we do not need KKT conditions or the Lagrangian. The disadvantage of direct constrained optimization is that it can make gradient descent fail to converge if the algorithm is not modified. Instead, we force the network to only use the most important $k$ locations in the feature map. The "importance" stems from an additional sparsity loss that forces the network to output an empty feature map. Part of the loss tries to increase $k$ locations, while another part tries to set all of them to zero. This constraint simplifies the optimization problem and allows us to keep the same accuracy as the unconstrained problem.

Restricting the output feature maps fundamentally changes the way the network works internally. In \cref{fig:mainfigure}, we see an example on how the constraint affects the class activation map (CAM). We also found that the networks trained with our approach become more robust. The intuition behind our proposed method is based on the observation that if the sample size of a class is too small, the network may tend to focus on the background instead of the object itself \cite{DBLP:journals/corr/abs-1911-08731, ribeiro2016why}. This can lead to undesirable biases in the classifier. In our approach, the constraint forces the network to not focus so much on the background.

The main contributions of this paper are:

\begin{itemize}
    \item \textbf{Size Priors:} We propose Top-GAP, a regularization technique incorporating a size prior directly into the network architecture. This method constrains the number of pixels the network utilizes during training and inference. It is beneficial for object classification tasks in contexts without perspective projections, such as biomedical imaging and datasets with centered objects.
    
    \item \textbf{Effective Receptive Field (ERF):} We link Top-GAP to the ERF and measure the influence of the feature output on the background and object pixels. We show that Top-GAP directs the network’s focus towards object pixels, reducing background influence.

    \item \textbf{Robustness to Adversarial Attacks and Distribution Shifts:} Further evidence that the background is less important is given by our robustness experiments. Top-GAP improves robustness against PGD and Square Attack, achieving up to 50\% higher accuracy without adversarial training. It also enhances accuracy by up to 5\% for datasets such as Waterbirds \cite{DBLP:journals/corr/abs-1911-08731}.

    \item \textbf{Intersection over Union:} By adjusting the pixel constraint, our method enables the network to focus more precisely on specific objects, leading to up to a 25\% improvement in Intersection over Union (IOU) compared to GradCAM and Recipro-CAM \cite{byun2023reciprocam}.
\end{itemize}

\section{Related Work}

Our work is related to different strands of research, each dealing with different aspects of improving the features and robustness of neural networks. This section outlines these research directions and introduces their relevance to our novel approach.

\textbf{Adversarial robustness.} It has been shown that neural networks are susceptible to small adversarial perturbations of the image \cite{goodfellow2015explaining}. For this reason, many methods have been developed to defend against such attacks. Some methods use additional synthetic data to improve robustness \cite{wang2023better, DBLP:journals/corr/abs-2110-09468}. \cite{wang2023better} makes use of diffusion models, while \cite{DBLP:journals/corr/abs-2110-09468} uses an external dataset. Other methods have shown that architectural decisions can influence robustness \cite{peng2023robust, huang2022exploring}. A disadvantage of all these approaches is that the clean accuracy and training speed are negatively affected \cite{DBLP:journals/corr/abs-1906-06032,clarysse2022adversarial}. 

\textbf{Bias mitigation and guided attention.} A notable line of research concentrates on channeling the network's focus towards specific feature subsets. Of concern is the prevalence of biases within classifiers, arising due to training on imbalanced data that perpetuates stereotypes \cite{pmlr-v81-buolamwini18a}. Biases may also stem from an insufficient number of samples \cite{burns2019women, zhao-etal-2017-men, bolukbasi2016man}, causing the network to emphasize incorrect features or leading to problematic associations. For instance, when the ground truth class is "boat", the network might focus on waves instead of the intended object.

\cite{he2023efficient, DBLP:journals/corr/abs-1905-00593} introduce training strategies to use CAMs as labels and refine the classifier's attention toward specific regions. In contrast, \cite{rajabi2022fair} proposes transforming the input images to mitigate biases tied to protected attributes like gender. Moreover, \cite{DBLP:journals/corr/abs-2104-14556} suggests a method to uncover latent biases within image datasets.

\textbf{Weakly-supervised semantic segmentation (WSSS).} \cite{DBLP:journals/corr/abs-1802-10171} focuses on accurate object segmentation given class labels. The Puzzle-CAM paper \cite{Jo_2021} introduces a novel training approach, which divides the image into tiles, enabling the network to concentrate on various segments of the object, enhancing segmentation performance. There are many more publications that focus on improving WSSS \cite{sun2023alternative}. Some making use of foundational models such as Segment Anything Model (SAM) \cite{kirillov2023segment} or using multi-modal models like CLIP \cite{radford2021learning}.

\textbf{Priors.} Prior knowledge is an important aspect for improving neural network predictions. For example, YOLOv2 \cite{redmon2016yolo9000} calculated the average width and height of bounding boxes on the dataset and forced the network to use these boxes as anchors. However, there are many other works that have tried to use some prior information to improve predictions \cite{8316924,9292601,9146293,7789620,pathak2015constrained}. In particular, \cite{pathak2015constrained} has proposed to add constraints during the training of the network. For example, they propose a background constraint to limit the number of non-object pixels. However, they only train the coarse output heat maps with convex-constrained optimization. The problem is that the use of constraints can make it harder to find the global optima. Therefore, it is harder to train the whole network.

\textbf{Our approach.} Much like bias mitigation strategies and attention-guided techniques, we direct the network's focus to specific areas. However, our approach does not require segmentation labels and only minimally changes the CNN architectures. The objective is to maintain comparable clean accuracy and the number of parameters, while significantly improving the interpretability of objects. In contrast to WSSS, we do not intend to segment entire objects, but instead continue to concentrate on the most discriminative features. Given that we modify the classification network itself, we also diverge from methods that solely attempt to enhance CAMs of pretrained models.

\section{Method}

In most cases of image classification, the majority of pixels are not important for the prediction. Usually, only a small object in the image determines the class. Our approach is geared towards these cases. In contrast, many modern CNNs implicitly operate under the assumption that every pixel in an image can be relevant for identifying the class. This perspective becomes evident when considering the global average pooling (GAP) layer \cite{lin2014network} used in modern CNNs. The aim of the GAP layer is to eliminate the width and height dimensions of the last feature matrix, thereby making it possible to apply a linear decision layer. 
The GAP layer averages all locations within the last feature matrix without making a distinction between the positions or values. This means that a corner position is treated in the same way as a center position. We also note that each of the locations in the last feature matrix corresponds to multiple pixels in the input image. This is known as the receptive field. Now, we want to define more formally the terminology.

\begin{definition}[Effective receptive field]\label{def:erf}
Let $X^{(p)}_{i^{(p)},j^{(p)}}$ be the feature matrix on the $p$th layer for $1 \leq p \leq n$ with coordinates $(i^{(p)}, j^{(p)})$. The input to the neural network is at $p = 1$ and the output feature map at $p = n$. Then the effective receptive field (ERF) of the output location $(i^{(n)}, j^{(n)})$ with respect to the input pixel $(i^{(1)}, j^{(1)})$ is given by $\frac{\partial X^{(n)}_{i^{(n)},j^{(n)}}}{\partial X^{(1)}_{i^{(1)},j^{(1)}}}$ \cite{luo2017understanding}.
\end{definition}

This definition assumes that each layer has only a single channel. For multiple output channels, we compute $\sum_{k=1}^{c^{(n)}}\frac{\partial X^{(n)}_{i^{(n)},j^{(n)},k}}{\partial X^{(1)}_{i^{(1)},j^{(1)}}}$ where $c^{(n)}$ are the channels of the last feature map. The ERF characterizes the impact of some input pixel on the output.

\begin{definition}[Global Average Pooling]
The feature output of the neural network $X^{(n)}$ is averaged to obtain a single value. This operation is known as Global Average Pooling (GAP) and is defined as:

$$\text{GAP}(X^{(n)}) = \frac{1}{h^{(n)}w^{(n)}} \sum_{i=1}^{h^{(n)}}\sum_{j=1}^{w^{(n)}} X^{(n)}_{i,j}\,,$$

\noindent where $h^{(n)}$ is the height and $w^{(n)}$ is the width of the output feature map. In practice, there is not only one channel but $c^{(n)}$ channels.
\end{definition}

An example shall explain the two terms. In case of EfficientNet-B0 \cite{tan2020efficientnet}, $X^{(n)}$ has dimension $7\times 7 \times 1280$ for an input image of size $224\times 224$ where $c^{(n)} = 1280$ are the channels. The $\text{GAP}(\cdot)$ operation reduces $X^{(n)}$ to a vector of size $1280\times 1$. All of the $7 \times 7$ locations have an effect on the classification. With the help of the ERF, we can measure how much the $224^2$ input pixels contribute to the $7^2$ output locations.

Another method to analyze what the neural network focuses on are the so-called class activation maps. These methods modify $X^{(n)}$ so that we get a visualization of what is important for the neural network.

\begin{definition}[Class Activation Map]
The product of multiplying the output tensor $X^{(n)}$ by some weight coefficient $W$ is known as a class activation map (CAM) \cite{zhou2015learning}. The standard CAM, also known as "CAM", uses the weights of the linear decision layer $L$.
\end{definition}

In the previous example, the linear decision layer $L$ would map the $1280$ channels to $c^{(n+1)}$ class channels. The output of the CAM would be in this case $7 \times 7 \times c^{(n+1)}$. Each of the $c^{(n+1)}$ maps can be upsampled to obtain a visualization.

\begin{definition}[GradCAM]
GradCAM is a generalization of CAM to non-fully convolutional neural networks (non-FCN) such as VGG. It is equivalent to the standard CAM for FCN like ResNet. It is defined as follows

$$\text{GradCAM}(X, c) = \text{ReLU}\left(\sum_{k} W_{k,c},X_{k}\right)\,,$$

\noindent with $W_{k,c} = \text{GAP}\left(\frac{\partial L(X)_c}{\partial X_k}\right)$, $k$ being the channel index of $X$ and $c$ being the index of the linear layer. Usually the last feature map $X^{(n)}$ is chosen for $X$.
\end{definition}

In addition to GradCAM, there are many other CAM methods. However, they are all based on reducing the channels of $X^{(n)}$ in order to obtain a visualization. Instead of improving GradCAM, as so many approaches have done before \cite{Chattopadhay_2018, DBLP:journals/corr/abs-1908-01224, 9462463, DBLP:journals/corr/abs-2008-02312, wang2020scorecam}, we propose that the output of the CNN should be both a CAM and a prediction. Then we can regularize the CAM during training and can more fundamentally influence what is highlighted in the CAM. 



Our approach involves integrating an object size constraint directly into the network, designed to enforce the utilization of a limited set of pixels for classification. This constraint allows for noise reduction and the elimination of unnecessary pixels from the CAM. In cases where specific-sized features determine the class, we can incorporate this prior knowledge into the neural network, enhancing its classification accuracy.

Before introducing the object constraint, we first change the model structure to output a higher-resolution CAM.

\subsection{Changing the model output structure} 
\begin{figure}[ht]
    \centering
    \includegraphics[scale=0.4]{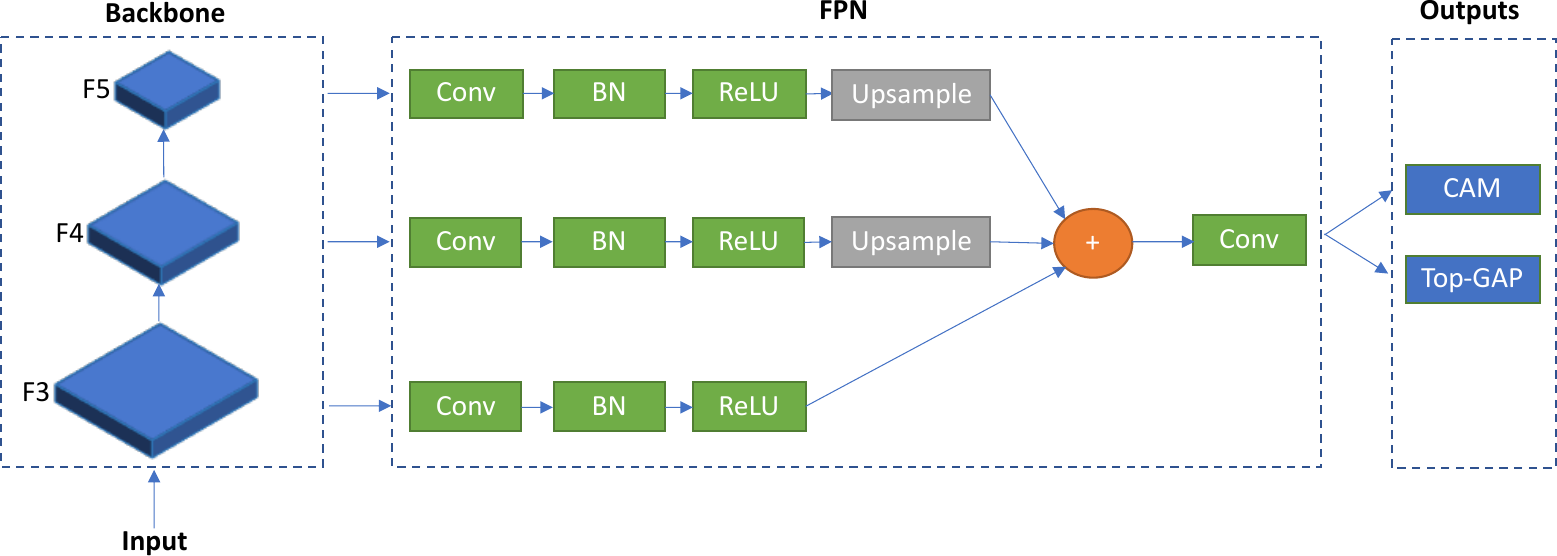}
    \caption{Example of our architecture applied to a backbone with 3 feature maps (e.g. $7\times 7$, $14\times 14$, $28\times 28$). For all convolutions except the final one, a kernel size of $3$ and $256$ filters is used. The last convolution employs a kernel size of $1$, with the number of filters set to match the number of output classes. The CAM is as large as the biggest feature map (here F3). Our pooling layer ("Top-GAP") averages the CAMs given by the last convolutional layer ("Conv") to create a vector containing the probability for each class. For the CAM, we disable "Top-GAP" and perform min-max scaling.}
    \label{fig:ourarch}
\end{figure}

\Cref{fig:ourarch} shows the general structure of our architecture. The backbone can be any standard CNN such as VGG \cite{simonyan2015deep}, ResNet \cite{he2015deep}, ConvNeXt \cite{liu2022convnet} or EfficientNet \cite{tan2020efficientnet}. Depending on the backbone, we use the last 3 or 4 feature maps as input to a feature pyramid network (FPN) \cite{lin2017feature}. We note that the original FPN as used for object detection was simplified in order to reduce parameters. All the feature maps are upsampled to the size of the largest feature map and added together. We found no advantage in using concatenation. This output is given to a final convolutional layer that has the number of output classes as filters. Note that a convolutional layer with kernel size 1 is used for the implementation of the final linear layer. Optionally, dropout can be applied as regularization during training. Lastly, we employ Top-GAP to obtain a single probability for each class. Top-GAP is introduced in the following section.

For convenience, we explicitly define two modes for our model (refer to \cref{fig:ourarch}):

\begin{enumerate}
    \item CAM: The output feature map is upsampled to the size of the input image and normalized to be in the range $[0, 1]$.
    \item training/prediction: Top-GAP is enabled to obtain the probabilities for each class.
\end{enumerate}

Without our modified model, we would need to use a method such as GradCAM to obtain a visualization.


Let us compare the two approaches: EfficientNet-B0 with GradCAM and EfficientNet-B0 with our output structure (see \cref{fig:ourarch}). GradCAM does not require any additional parameters because it generates the activation map from the model itself. If we change the model structure, we have more parameters, but also more influence on what is seen in the CAM. If we were to replace GradCAM with LayerCAM or some other method, it would never have the same impact as changing the model training itself (our approach). In addition, GradCAM does not combine multiple feature maps by default to achieve better localization.

In our approach, the standard output linear layer of some classification model like EfficientNet-B0 is substituted with $f+1$ convolutional layers, where $f$ corresponds to the number of feature maps. This leads to a small increase in the number of parameters. 

\begin{wraptable}{l}{0.7\textwidth}
  \centering
\scalebox{0.9}{
  \begin{tabular}{lll}
    \toprule
    Architecture & Params (unmodified) & Params (ours) \\
    \midrule
    VGG11-BN       & 132.87M & 12.43M \\
    EfficientNet-B0 & 4.08M & 4.75M \\
    DenseNet-121   & 7.98M  & 8.03M  \\
    \bottomrule
  \end{tabular}}
  \caption{Number of parameters for some architectures. We have fewer parameters than VGG because all additional linear layers are removed.}
  \label{tab:numparams}
\end{wraptable}

As indicated in \cref{tab:numparams}, we can achieve a comparable number of parameters.

These changes to the model are prerequisites for enabling the integration of size constraints within the neural network. If only the last feature map were used, a single value would correspond to an excessively large area in the original image. Hence, combining multiple feature maps proves advantageous. This idea is reinforced by findings from \cite{9462463}, which highlight that employing multiple layers enhances the localization capabilities of CAMs.

\subsection{Defining the pixel constraint (Top-GAP)} Instead of using the standard GAP layer, we replace the average pooling by a top-k pooling, where only the $k$ highest values of the feature matrix are considered for averaging. This pooling layer limits the number of input pixels that the network can use for generating predictions.

In a standard CNN, the last feature map is at layer $n$. In our model (\cref{fig:ourarch}), the last feature map is at $n+1$ because we replaced the linear decision layer $L$ by a $1\times 1$ convolution.

\begin{definition}[Top-GAP] We define the Top-GAP layer as follows:

$$\text{Top-GAP}(\tilde{X}, k)_t = \frac{1}{k} \sum_{i=1}^k \tilde{X}_{i,t}\,,$$

where $\tilde{X}$ represents the ordered feature matrix $X^{(n+1)}$ with dimensions\\
$h^{(n+1)}w^{(n+1)} \times c^{(n+1)}$, where $c^{(n+1)}$ corresponds to the number of output classes. Each of the $c^{(n+1)}$ column vectors is arranged in descending order by value, and $k$ values are selected. $i$ indicates the ranking, with $i=1$ being the largest value and $i=k$ being the smallest. $t$ is an index indicating the channel. We select for each channel different values.
\end{definition}

When $k = 1$, we obtain global max pooling (GMP). When $k = h^{(n+1)}w^{(n+1)}$, the layer returns to standard GAP. The parameter $k$ enforces the pixel constraint, and its value depends on the image size. For instance, if the largest feature map has dimensions $56\times 56$, then $\frac{k}{56^2}$ values are selected. Hence, when adjusting this parameter, it is crucial to consider the relative object size in the highest feature map.

\subsection{Classification loss function} The last component of our method involves changing the loss function. While the $\text{Top-GAP}(\cdot)$ layer considers only locations with the highest values, these locations might not necessarily be the most important ones. Thus, it becomes essential to incentivize the reduction of less important positions to zero.

To achieve this, we add an $\ell_1$ regularization term to the loss function, inducing sparsity in the output. The updated loss function is defined as follows:

\begin{equation}
L = \lambda||X^{(n+1)}||_1 + \text{CE}\left(\hat{y}, y\right)\,,
\label{eq:2}
\end{equation}

where $\text{CE}(\hat{y}, y)$ represents the cross-entropy loss between the prediction $\hat{y} = \text{softmax}\left(\text{Top-GAP}(\tilde{X}, k)\right)$ and the ground truth $y$. $\tilde{X}$ is the ordered $X^{(n+1)}$ feature output in our model, while $k$ is a fixed non-trainable parameter. Here, $\lambda$ controls the strength of the regularization. We found that for most datasets $\lambda = 1$ is sufficient.

\section{Evaluation}

In this section, we will systematically test the claims of our method on several datasets. Since there is no ground truth for explainability, we focus mainly on surrogate measures. Our main surrogate measure for interpretability is the background of the image. We show that our method causes the network to focus less on it. Furthermore, we also evaluate how our model behaves in the presence of distribution shifts. A description of the datasets used here can be found in the appendix \ref{descriptiondataset}.

Our method consists of three components: Top-GAP, model structure, and loss function. Top-GAP has a hyperparameter $k$ that defines the number of input pixels the network should use. We tested values $k \in \{64, 128, \dots, 1024, 2048\}$ and only report the result that maximizes the metric (e.g. accuracy).

\subsection{Hypothesis: the gradient of object pixels becomes more important}

We want to show that with our method not all pixels in the input image have the same influence on the output feature map $X^{(n+1)}$. Recall that in our model, the last feature map is at $n+1$ because we have replaced the linear layer with a convolutional layer.

In most datasets (e.g. ImageNet), the object to be classified is located in the center of the input image. While each pixel in the input image corresponds to multiple values in the output feature map $X^{(n+1)}$, the general position is the same. The center in the output is also the center in the input.

The input pixels should contribute much more to the center than to the background of $X^{(n+1)}$. We want to quantify how much influence the input pixels have on the center of $X^{(n+1)}$ and on the corner of $X^{(n+1)}$. For this, we use \cref{def:erf} and define a metric.

\begin{definition}[ERF distance]
Let $\text{ERF}(1,1) = \frac{1}{hw}\sum_{i,j}\left|\frac{\partial X^{(n+1)}_{1,1}}{\partial X^{(1)}_{i,j}}\right|$ to be the absolute change of the output corner position $(1,1)$ with all input pixels $(i, j)$. Similarly, we define $\text{ERF}(\frac{h}{2},\frac{w}{2})$ to be the change of the output center position with respect to the input, where $h$ and $w$ is the width of the output feature map. Then the ERF distance is $\text{ERF}(\frac{h}{2},\frac{w}{2}) - \text{ERF}(1,1)$.
\end{definition}

Intuitively, we expect a low value for $\text{ERF}(1,1)$ because the corner position of the feature map contains less information. Similarly, $\text{ERF}(\frac{h}{2},\frac{w}{2})$ should be a high value because the object is in the center. If the difference between the two values is low, it means that each pixel contributes similarly to the output.

\Cref{tab:erf} shows that for the standard CNN the center of the image has the same effect as the corner. $\text{ERF}(1,1)$ has the same value range as $\text{ERF}(\frac{h}{2},\frac{w}{2})$. Compare this to our approach, where there is a large difference between the center and the corner ERF.

During backpropagation, the neural network goes from the end to the beginning of the network and updates the weights. With our method, we set the gradient at the background positions of the last feature matrix to almost zero. This also affects all other layers as a consequence of the chain rule.

\begin{table}[h!]
  \centering
\scalebox{0.9}{
  \begin{tabular}{llcc}
    \toprule
    Dataset & Arch & ERF distance $\uparrow$ & ERF distance (ours) $\uparrow$ \\
    \midrule
    \multirow{3}{*}{COCO \cite{lin2015microsoft}} & EN & 0.108 & \textbf{0.447} \\
    & CN & 0.072 & \textbf{0.288} \\
    & RN & 0.273 & \textbf{0.399} \\
    \midrule
    \multirow{2}{*}{Oxford \cite{parkhi12a}} & EN & 0.013 & \textbf{0.383} \\
    & RN & 0.060 & \textbf{0.443} \\
    \midrule
    \multirow{3}{*}{CUB-200-2011 \cite{wah_branson_welinder_perona_belongie_2011}} & EN & -0.033 & \textbf{0.480} \\
    & CN & -0.034 & \textbf{0.242} \\
    & RN & 0.092 & \textbf{0.529} \\
    \bottomrule
  \end{tabular}}
  \caption{The table shows that our approach leads to a different ERF. The center has a stronger effect than the corner of the image. "Ours" is our approach (with pixel constraint, $\ell_1$ loss and the changes to the model). The other column is the standard model without any changes. EN = EfficientNet-B0, CN = ConvNeXt-tiny, RN = ResNet-18.}
  \label{tab:erf}
\end{table}

\begin{figure}[h!]
    \centering
    \begin{subfigure}{0.4\textwidth}
        \centering
        \includegraphics[width=\textwidth]{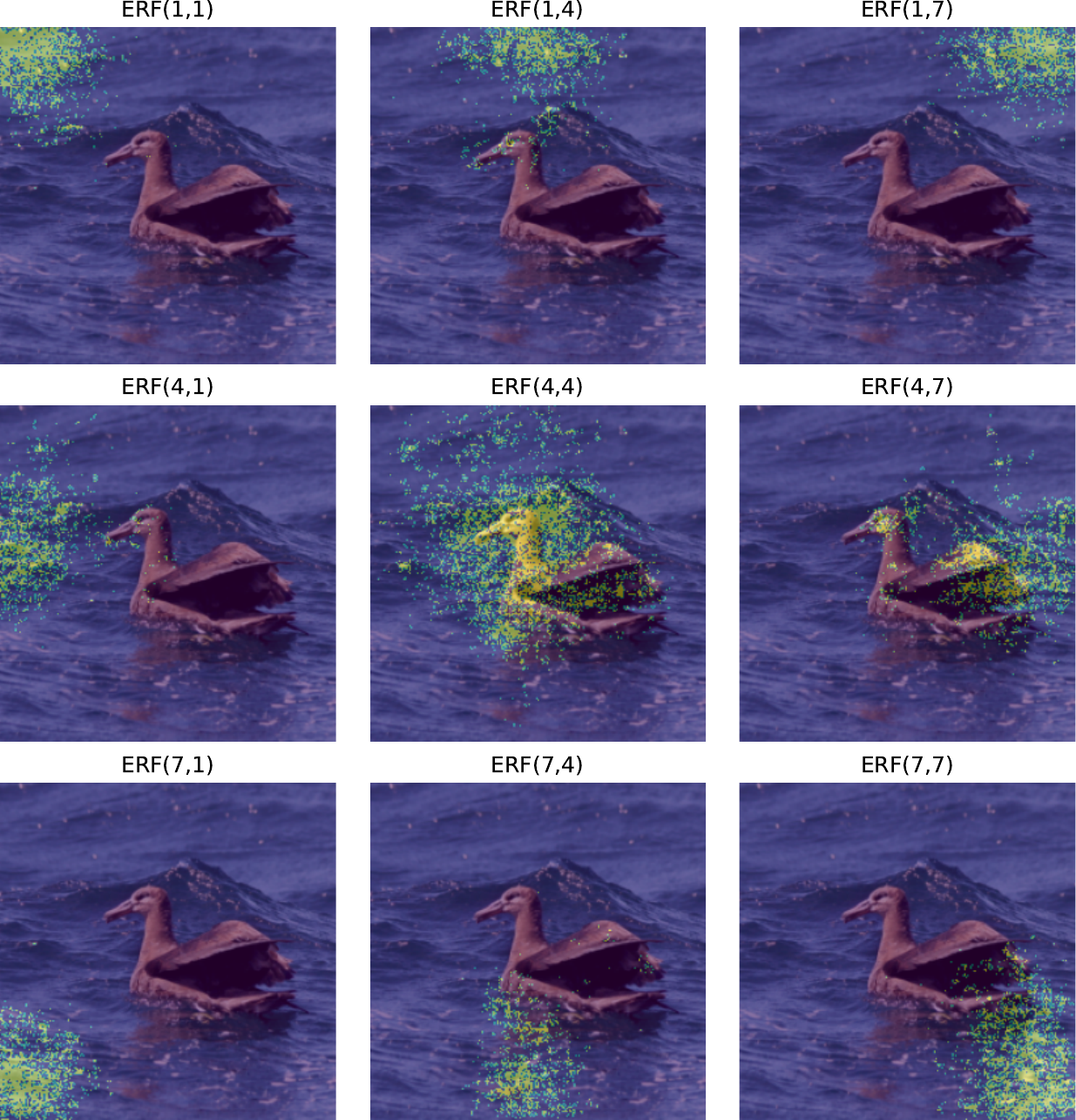}
        \caption{Standard ResNet-18}
        \label{fig:subfig2a}
    \end{subfigure}
    \hspace{1cm}
    \begin{subfigure}{0.4\textwidth}
        \centering
        \includegraphics[width=\textwidth]{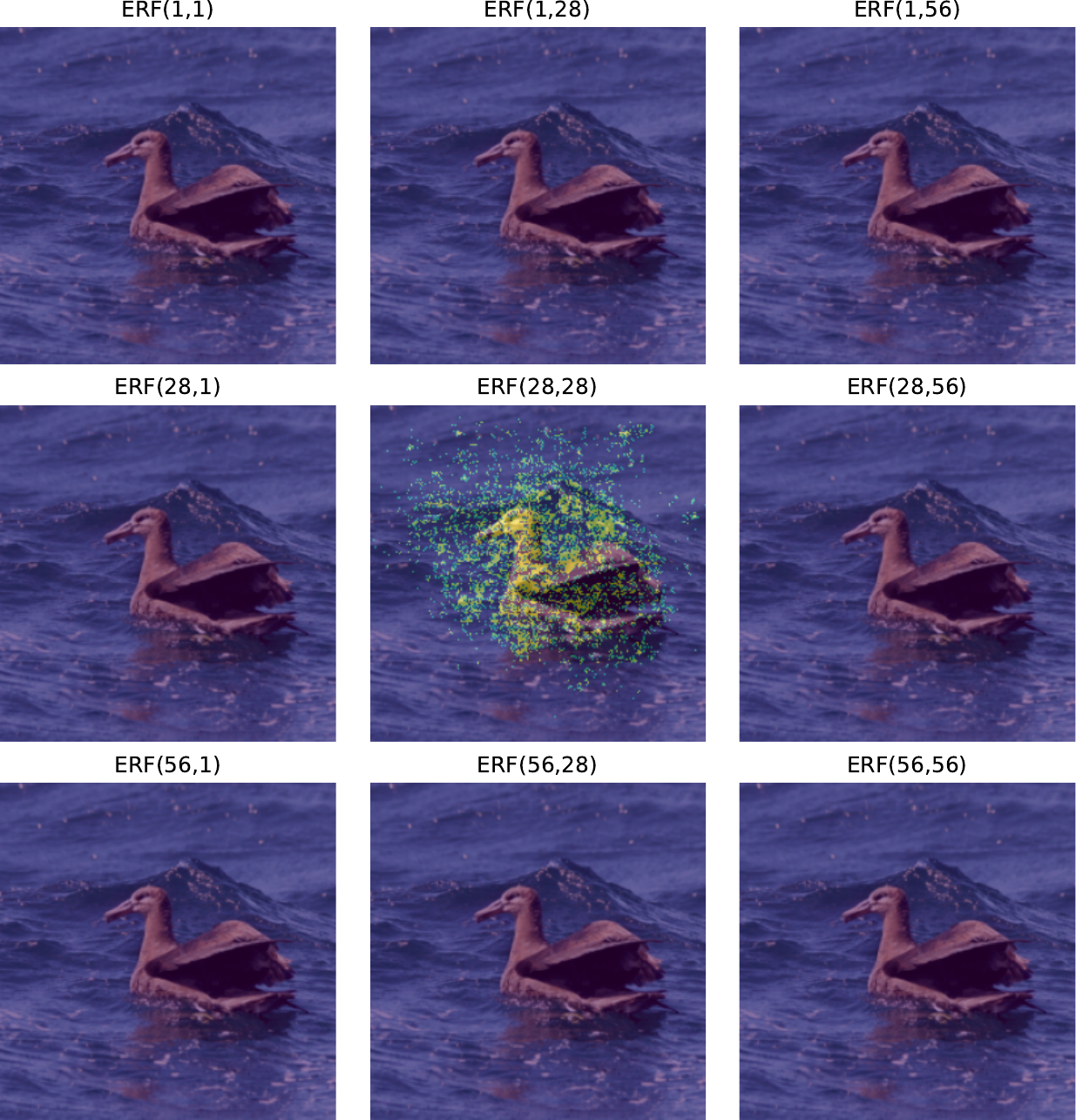}
        \caption{ResNet-18 with our approach}
        \label{fig:subfig2b}
    \end{subfigure}
    \caption{ERF for various locations in the output feature map. The background becomes less important using our approach. The last feature map of standard ResNet has size $7\times 7$, with our approach it has size $56\times 56$.}
    \label{fig:erfexperiment}
\end{figure}

The visualization in \cref{fig:erfexperiment} confirms the numerical results. Since there are $7^2 = 49$ positions for the standard ResNet and $56^2$, we only considered $9$ pixel positions. We see that the gradient disappears at the locations where there is no object. More numerical details are provided in the appendix in \cref{tab:erfdetails1} and \cref{tab:erfdetails2}.

\subsection{Hypothesis: the background is less susceptible to adversarial attacks}

The last experiment showed that by changing the values in the output feature matrix, we also change the effect of the input pixels. Another idea to prove that the network focuses less on the background is to use adversarial attacks. The goal of such an attack is to change the input pixels so that the classification prediction is different. Using our method, we expect the attack area to be smaller since the network uses the background less.

It is important to emphasize that we only need to achieve higher robustness against standard models. Our goal here is to force the network to focus on different regions in order to achieve better interpretability. We do not intend to compete with adversarially trained networks. Adversarial training (AT) is slower, leads to less clean accuracy and does not make the networks more interpretable.

\begin{table}[h!]
  \centering
  \scalebox{0.9}{\begin{tabular}{llllllll}
        \toprule
        Method & Arch & PGD$^{20}$ $\uparrow$ & PGD$^{50}$ $\uparrow$ & Square $\uparrow$ & Clean $\uparrow$\\\hline
Standard & PRN18 & 0.0 & 0.0 & 0.0 & 0.945\\
\midrule
Top-GAP (ours) & PRN18 & 0.517 & 0.313 & 0.343 & \textbf{0.951}\\
\midrule
FGSM-AT \cite{andriushchenko2020understanding} & PRN18 & - & \textbf{0.476} & - & 0.81\\ 
SAT \cite{peng2023robust} & RN50 & \textbf{0.552} & - & - & 0.849\\
        \bottomrule
        \end{tabular}}
  \caption{Results on CIFAR-10. We use $\ell_{\infty} = \nicefrac{8}{255}$ and 20/50 steps (PGD). For AT models, we report the values from the papers. SAT = Standard Adversarial Training, PRN18 = PreAct ResNet-18, RN50 = ResNet-50. Our results are close to the robustness of adversarially trained networks.}
  \label{tab:robustness2}
\end{table}

\cref{tab:robustness2} shows that we outperform the standard models by far and even achieve comparable clean accuracy. Notably, square attack \cite{DBLP:conf/eccv/AndriushchenkoC20} does not rely on local gradient information. It should, therefore, be not affected by gradient masking. This shows that our robustness is not necessarily a result of "shattered gradients" \cite{athalye2018obfuscated}. On other datasets, such as ImageNet, we similarly see small increases in robustness while keeping the same accuracy (refer to the appendix \cref{tab:adv_attacks_results}).

To make the argument that the network is less susceptible to attacks to the background even more convincing, we consider the FGSM attack \cite{https://doi.org/10.48550/arxiv.1412.6572 }. It uniformly perturbs each pixel by $\pm k$ for some $1 \leq k \leq 255$. Instead of perturbing the whole image, we perturb either just the object or just the background. For this, we use the segmentation mask provided by the CUB dataset.

\begin{definition}[Attack distance]
Let $\text{SAR}(I) = 1 - \operatorname{Acc}$ be the successful attack rate, given perturbed images $I$. We define $\text{SAR}(O) - \text{SAR}(B)$ to be the attack distance (AD) between the object image $O$ and the background image $B$.
\end{definition}

In image $O$, only the pixels of the object were changed by $\pm 1$, while all other pixels of the original image were retained. Similarly, in image $B$, only the background was perturbed while the object remained untouched. Just like the ERF distance, we expect $\text{SAR}(O)$ to be large and $\text{SAR}(B)$ to be small. FGSM should be more successful if it attacks the object and less successful if it attacks the background.

\begin{wraptable}[9]{l}{0.5\textwidth}
\vspace{-10pt}  
\centering
\scalebox{0.8}{\begin{tabular}{lll}
\toprule
Arch & Method    & Attack distance \\
\midrule
\multirow{2}{*}{EN} & Standard          & 0.016 \\
                    & ours      & \textbf{0.064} \\
\cmidrule{1-3}
\multirow{2}{*}{RN} & Standard          & 0.022 \\
                    & ours     & \textbf{0.065} \\
\cmidrule{1-3}
\multirow{2}{*}{CN} & Standard          & 0.078 \\
                    & ours      & \textbf{0.132} \\
\bottomrule
\end{tabular}}
\caption{The background is less susceptible to attacks with our approach. The dataset is CUB-200-2011.}
\label{tab:ad}
\end{wraptable}

In \cref{tab:ad}, we see that when using the standard networks, the values of $\text{SAR}(O)$ and $\text{SAR}(B)$ are close to each other. This means that the center has the same effect as the background. The network concentrates on all pixels equally. With our approach, we can manipulate the class more easily by changing the object pixels.

Background pixels, on the other hand, are less important. For this reason, we have a higher AD value. This gives a second proof for our hypothesis.

\subsection{Hypothesis: classification makes use of object pixels}

Another way to show that we are directing the attention of the network to the object is through distribution shifts. To address this, we use the Waterbirds dataset \cite{DBLP:journals/corr/abs-1911-08731}, where the backgrounds of images are replaced. Furthermore, we evaluate accuracy on ImageNet-Sketch \cite{DBLP:journals/corr/abs-1905-13549} and ImageNet-C \cite{DBLP:journals/corr/abs-1903-12261}.

\begin{table}[h!]
  \centering
\scalebox{0.8}{
  \begin{tabular}{llcc}
    \toprule
    Dataset & Arch & Acc $\uparrow$ & Acc $\uparrow$ (ours) \\
    \midrule
    \multirow{3}{*}{CUB $\to$ Waterbirds} & EN & 0.521 & \textbf{0.564} \\
    & CN & 0.722 & \textbf{0.737} \\
    & RN & 0.468 & \textbf{0.520} \\
    \midrule
    \multirow{2}{*}{ImageNet $\to$ Sketch} & VG & 0.179 & \textbf{0.200} \\
    & RN & 0.206 & \textbf{0.236} \\
    \midrule
    \multirow{2}{*}{ImageNet $\to$ ImageNet-C} & VG & 0.494 & \textbf{0.498} \\
    & RN & 0.513 & \textbf{0.535} \\
    \bottomrule
  \end{tabular}}
  \caption{Evaluation of the out-of-distribution accuracy by using images outside the original dataset. $X \to Y$ means train on X and validate on Y.}
  \label{tab:distshift}
\end{table}

An improvement in accuracy can be observed for all datasets. While there are works that show higher accuracy for datasets such as ImageNet-Sketch \cite{fang2022eva}, they are based on specialized training methods (self-supervised, semi-supervised) and/or more data. Our proposed method comes "without cost" in the sense that it works for any architecture and dataset, without requiring more GPU resources. It can be viewed as a regularization technique. This increase in robustness does not negatively affect the accuracy. We also see a comparable accuracy when using 5-fold stratified cross validation (refer to \cref{tab:acctable}).

\begin{table}[ht]
  \centering
\scalebox{0.8}{
\begin{tabular}{llcc}
        \toprule
        Dataset & Arch & Accuracy $\uparrow$ & Accuracy (ours) $\uparrow$\\
        \midrule
        \multirow{3}{*}{COCO} & EN & 0.801 $\pm$ 0.009 & \textbf{0.803} $\pm$ 0.006\\
         & CN & 0.939 $\pm$ 0.006 & \textbf{0.940} $\pm$ 0.005\\
         & RN & 0.853 $\pm$ 0.004 & \textbf{0.868} $\pm$ 0.005\\
        \midrule
        \multirow{2}{*}{Wood} & EN & 0.672 $\pm$ 0.037 & \textbf{0.681} $\pm$ 0.041\\
         & CN & 0.721 $\pm$ 0.030 & \textbf{0.724} $\pm$ 0.033\\
        \midrule
        \multirow{2}{*}{Oxford} & EN & 0.854 $\pm$ 0.008 & \textbf{0.863} $\pm$ 0.010\\
         & RN & 0.861 $\pm$ 0.007 & \textbf{0.862} $\pm$ 0.007\\
        \midrule
        \multirow{3}{*}{CUB} & EN & 0.76 $\pm$ 0.01 & \textbf{0.77} $\pm$ 0.005\\
         & RN & \textbf{0.69} $\pm$ 0.014 & 0.685 $\pm$ 0.006\\
         & CN & \textbf{0.862} $\pm$ 0.007 & 0.854 $\pm$ 0.005\\
        \midrule
        \multirow{2}{*}{ImageNet} & VG & \textbf{0.704} & 0.699\\
         & RN & \textbf{0.698} & 0.697\\
        \bottomrule
\end{tabular}}
  \caption{Our approach refers to the changed model with pixel constraint and $\ell_1$ loss. The original models come from PyTorch Image Models \cite{rw2019timm} and are pretrained on ImageNet. EN = EfficientNet-B0, CN = ConvNeXt-tiny, RN = ResNet-18, VG = VGG11-bn. For ImageNet, we only use a train/val split.}
  \label{tab:acctable}
\end{table}

\subsection{Hypothesis: increased interpretability due to pixel constraints}

\begin{wrapfigure}[12]{r}{0.6\textwidth}

\vspace{-25pt}  
    \centering
    \includegraphics[scale=0.35]{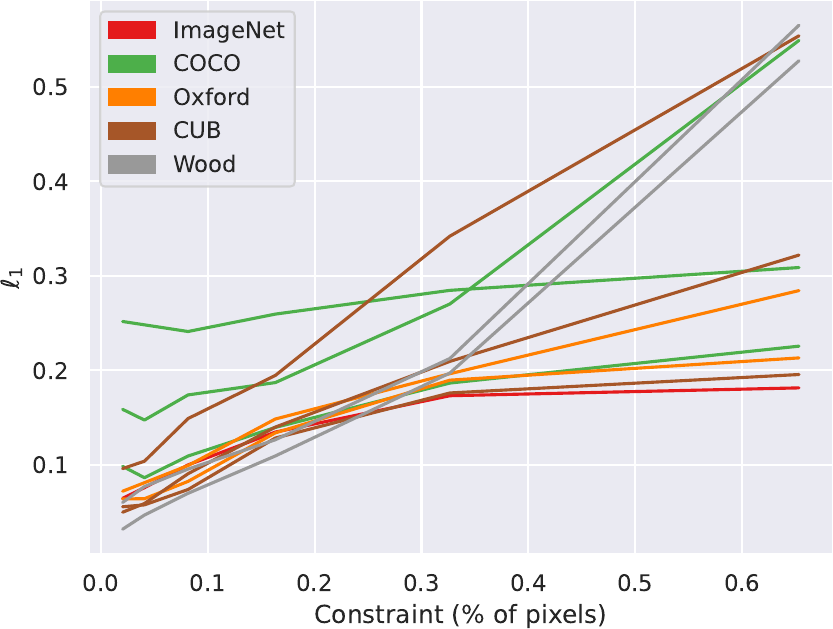}
    \caption{Each line in the graph represents a dataset+architecture combination. The x-axis shows the normalized $k$ value (e.g. $\frac{64}{56^2}$) for the constraint, while the y-axis represents the $\ell_1$ norm.}
    \label{fig:trendconstraint}
\end{wrapfigure}

The last experiments have shown that we can direct the attention of the network to the object. So far, we have used the pixel constraint $k$, which maximizes the metric. However, it is also possible to vary this value to incorporate human knowledge into the prediction. In \cref{fig:trendconstraint}, we measure the sparsity of our CAM.

It is evident that as we increase the constraint $k$, the number of displayed pixels in the CAM also rises. More experiments are provided in \cref{tab:l1} and \cref{sec:detailsl1} in the appendix.

\begin{figure}[ht]
    \centering
        \includegraphics[scale=0.15]{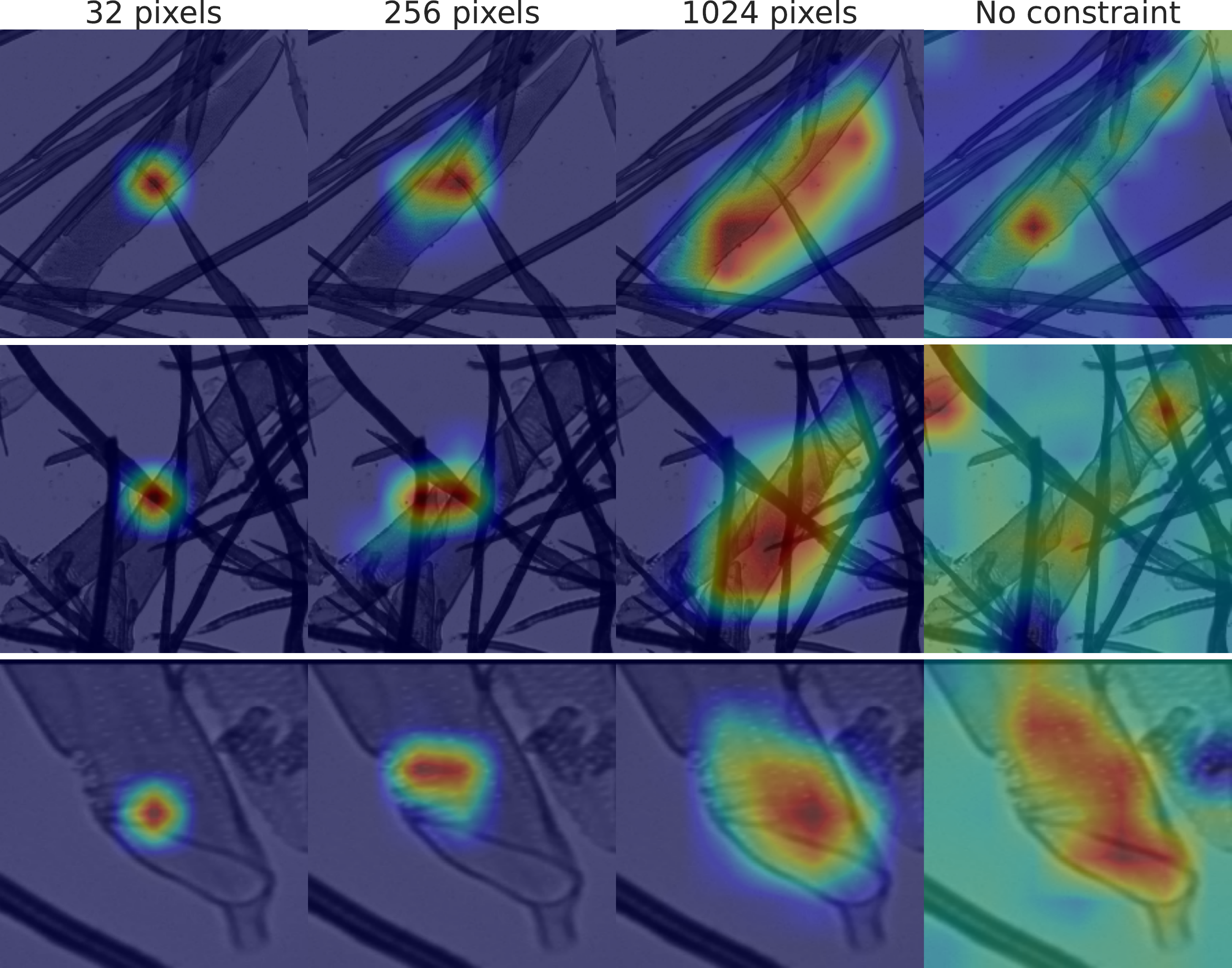}
    \caption{Impact of pixel constraint on CAM (Wood identification dataset \cite{nieradzik2023automating}). "No constraint" denotes a standard unmodified EfficientNet-B0 model using CAM/GradCAM \cite{Selvaraju_2019}. The object in the center, known as a vessel should be highlighted. Without our method, the background containing fibers is also highlighted.}
    \label{fig:woodexample}
\end{figure}

We evaluated our approach on a real-world dataset with microscopic images. \cref{fig:woodexample} shows that we can use our constraint to direct the focus of the network to the vessels. The standard model focuses on the background or fibers instead. For biologists, however, only the vessels are important.

Finally, we assess the segmentation overlap. While segmentation masks should not be considered as ground truth for explainability, they provide valuable insights into the focus of the network. This analysis allows us to quantitatively measure whether the network is predominantly focused on the object of interest or on the surrounding background.

\begin{table}[ht]
  \centering
\scalebox{0.9}{
  \begin{tabular}{llccc}
    \toprule
    Dataset & Arch & IOU (GC) $\uparrow$ & IOU (RC) $\uparrow$ & IOU (ours) $\uparrow$ \\
    \midrule
    \multirow{3}{*}{COCO} & EN & 0.309 & 0.245 & \textbf{0.348} \\
    & CN & 0.103 & 0.268 & \textbf{0.361} \\
    & RN & 0.371 & 0.359 & \textbf{0.391} \\
    \midrule
    \multirow{3}{*}{CUB} & EN & 0.323 & 0.337 & \textbf{0.414} \\
    & CN & 0.125 & 0.279 & \textbf{0.389} \\
    & RN & 0.268 & 0.34 & \textbf{0.435} \\
    \bottomrule
  \end{tabular}}
  \caption{Comparison of Intersection over Union (IOU) scores across different methods and architectures. The IOU (GC) column represents the standard unchanged model using GradCAM (GC). Similarly, RC is Recipro-CAM  \cite{byun2023reciprocam}.}
  \label{tab:ioutable}
\end{table}

\cref{tab:ioutable} gives even more evidence that the pixel constraint allows the network to focus more on the object.

\section{Discussion and Outlook}

In this paper, we presented a new approach to improve the explainability of CNNs. Our method focuses on controlling the number of pixels a network can use for predictions, resulting in CAMs with lower noise and better localization. The results show that our approach is effective on a variety of datasets and architectures. We have consistently observed both visually and numerically more concise feature representations in the CAMs. In addition, our approach provides a novel form of network regularization. By forcing the network to focus exclusively on objects of a predefined size, we reduce the risk of highlighting irrelevant regions, which can be critical for applications that require precise object localization or for reducing bias.

\textbf{Limitations.} Determining the optimal value for the pixel constraint parameter $k$ currently depends on hyperparameter tuning. It is possible to explore automated methods for determining this parameter to improve efficiency and adaptability. Second, given the variety of object sizes, it may not be ideal to rely on a single parameter for all objects. Only in specific areas such as biomedical imaging, where object size are not influenced by perspective projections (e.g. microscope) typically show low size variances. Investigating ways to dynamically adjust this parameter for different object sizes would be a valuable line of research. Finally, the proposed FPN module can be further refined to improve accuracy even more.


%
%
\bibliographystyle{splncs04}
\bibliography{main}

\appendix

\include{appendix}

\end{document}

%% file: appendix.tex
\clearpage

\onecolumn
\appendix
\setcounter{figure}{0}
\renewcommand{\thefigure}{A\arabic{figure}}
\setcounter{table}{0}
\renewcommand{\thetable}{A\arabic{table}}

\setcounter{page}{1}
{
    \centering
    \Large
    \textbf{\MyTitle} \\
    \vspace{0.5em}Supplementary Material \\
    \vspace{1.0em}
}
The appendix contains the following additional materials:

\begin{itemize}
\item A detailed description of the datasets: \ref{descriptiondataset}.
\item More details regarding the effective receptive field: \cref{tab:erfdetails1}, \cref{tab:erfdetails2}.
\item More details regarding adversarial attacks: \cref{tab:adv_attacks_results}.
\item More details regarding sparsity: \cref{tab:l1}, \cref{tab:tradeoffl1}, \cref{tab:tradeoffl12}, \cref{tab:tradeoffl13}.
\end{itemize}

\section{Description of datasets}
\label{descriptiondataset}

We test all our models on the following datasets:

\begin{itemize}
    \item COCO \cite{lin2015microsoft}: We turned this segmentation dataset into a classification dataset by excluding images with more than one object. Furthermore, we kept only classes with a minimum of 20 samples per class. The resulting subset comprises 53 classes.
    \item Wood identification dataset \cite{nieradzik2023automating}: This dataset consists of high-resolution microscopy images for hardwood fiber material. Nine distinct wood species have to be distinguished.
    \item Oxford-IIIT Pet Dataset \cite{parkhi12a}: The task is to differentiate among 37 breeds of dogs and cats.
    \item CUB-200-2011 \cite{wah_branson_welinder_perona_belongie_2011} and Waterbirds \cite{DBLP:journals/corr/abs-1911-08731}: 200 classes of birds have to be distinguished. Waterbirds replaces the background of the original images to test the models for biases. 
    \item ImageNet \cite{5206848}: A large-scale dataset with 1000 different classes. ImageNet-Sketch \cite{DBLP:journals/corr/abs-1905-13549} / ImageNet-C \cite{DBLP:journals/corr/abs-1903-12261} replaces the original validation images with out-of-distribution / corrupted images.
    \item CIFAR10: A dataset where each image has a size of $32\times 32$. 10 classes have to be distinguished.
\end{itemize}

\clearpage

\section{Effective Receptive Field (ERF)}

In \cref{tab:erf}, we only showed the difference between the center and the corner ERF. In the following tables, we provide the individual values. The gradients were z-normalized to have mean at $0$ and standard deviation at $1$.

\begin{table}[ht]
  \centering
\scalebox{0.9}{
\begin{tabular}{lllll}
    \toprule
    Dataset & Architecture & Center ERF $\uparrow$ & Center ERF (ours) $\uparrow$ \\
    \midrule
    \multirow{3}{*}{COCO \cite{lin2015microsoft}} 
        & EfficientNet-B0 & 0.534 & 0.54 \\
        & ConvNeXt-tiny & 0.47 & 0.439 \\
        & ResNet-18 & 0.595 & 0.571 \\
    \midrule
    \multirow{2}{*}{Oxford \cite{parkhi12a}} 
        & EfficientNet-B0 & 0.087 & 0.51 \\
        & ResNet-18 & 0.104 & 0.542 \\
    \midrule
    \multirow{3}{*}{CUB-200-2011 \cite{wah_branson_welinder_perona_belongie_2011}} 
        & EfficientNet-B0 & 0.489 & 0.493 \\
        & ConvNeXt-tiny & 0.477 & 0.398 \\
        & ResNet-18 & 0.538 & 0.534 \\
    \midrule
    Average & - & 0.412 & \textbf{0.503} \\
    \bottomrule
\end{tabular}}
\caption{Center ERF. "Ours" is our approach (with pixel constraint, $\ell_1$ loss, and changes to the model). The other columns are the standard models without any changes.}
\label{tab:erfdetails1}
\end{table}

The table shows that for the center pixel the gradient with respect to the input image is higher, when using our method.
\begin{table}[ht]
  \centering
\scalebox{0.9}{
\begin{tabular}{llll}
    \toprule
    Dataset & Architecture & Corner ERF $\downarrow$ & Corner ERF (ours) $\downarrow$ \\
    \midrule
    \multirow{3}{*}{COCO \cite{lin2015microsoft}} 
        & EfficientNet-B0 & 0.426 & 0.093 \\
        & ConvNeXt-tiny & 0.398 & 0.151 \\
        & ResNet-18 & 0.322 & 0.172 \\
    \midrule
    \multirow{2}{*}{Oxford \cite{parkhi12a}} 
        & EfficientNet-B0 & 0.074 & 0.127 \\
        & ResNet-18 & 0.044 & 0.099 \\
    \midrule
    \multirow{3}{*}{CUB-200-2011 \cite{wah_branson_welinder_perona_belongie_2011}} 
        & EfficientNet-B0 & 0.522 & 0.013 \\
        & ConvNeXt-tiny & 0.511 & 0.156 \\
        & ResNet-18 & 0.446 & 0.005 \\
    \midrule
    Average & - & 0.343 & \textbf{0.102} \\
    \bottomrule
\end{tabular}}
\caption{Corner ERF. The values are lower using our approach, indicating improved performance. EN = EfficientNet-B0, CN = ConvNeXt-tiny, RN = ResNet-18.}
\label{tab:erfdetails2}
\end{table}

Similarly, we see that the pixels have less of an effect when the corner of the input image is considered.

\clearpage

\section{Adversarial Attacks}

We evaluated our approach using adversarial attacks on various datasets. For all the experiments summarized in \cref{tab:adv_attacks_results}, we use an $\ell_{\infty}$ constraint of $\nicefrac{1}{255}$ for both FGSM and PGD attacks. The PGD attacks were performed with 40 steps. The $\pm$ symbol denotes the standard deviation of the accuracy across 5 different folds. Due to computational complexity, we report only a single run for the ImageNet dataset.

\begin{table}[ht]
  \centering
  \scalebox{0.7}{
  \begin{tabular}{llcccccc}
    \toprule
    Dataset & Architecture & FGSM $\uparrow$ & FGSM $\uparrow$ (ours) & PGD $\uparrow$ & PGD $\uparrow$ (ours) & Clean Acc & Clean Acc (ours) \\
    \midrule
    \multirow{3}{*}{COCO} & EfficientNet-B0 & 0.07 & \textbf{0.3063} & 0.0 & \textbf{0.1098} & 0.801 $\pm$ 0.009 & \textbf{0.803} $\pm$ 0.006 \\
                          & ConvNeXt-tiny & 0.51 & \textbf{0.678} & 0.301 & \textbf{0.463} & 0.939 $\pm$ 0.006 & \textbf{0.940} $\pm$ 0.005 \\
                          & ResNet-18 & 0.288 & \textbf{0.394} & 0.08 & \textbf{0.142} & 0.853 $\pm$ 0.004 & \textbf{0.868} $\pm$ 0.005 \\
    \midrule
    \multirow{2}{*}{Wood} & EfficientNet-B0 & 0.0 & \textbf{0.277} & 0.0 & \textbf{0.085} & 0.672 $\pm$ 0.037 & \textbf{0.681} $\pm$ 0.041 \\
                          & ConvNeXt-tiny & 0.0 & \textbf{0.404} & 0.0 & \textbf{0.01} & 0.721 $\pm$ 0.030 & \textbf{0.724} $\pm$ 0.033 \\
    \midrule
    \multirow{2}{*}{Oxford} & EfficientNet-B0 & 0.037 & \textbf{0.107} & \textbf{0.0} & \textbf{0.0} & 0.854 $\pm$ 0.008 & \textbf{0.863} $\pm$ 0.010 \\
                            & ResNet-18 & 0.104 & \textbf{0.281} & 0.016 & \textbf{0.104} & 0.861 $\pm$ 0.007 & \textbf{0.862} $\pm$ 0.007 \\
    \midrule
    \multirow{3}{*}{CUB} & EfficientNet-B0 & 0.04 & \textbf{0.147} & 0.0 & \textbf{0.04} & 0.76 $\pm$ 0.01 & \textbf{0.77} $\pm$ 0.005 \\
                         & ResNet-18 & 0.06 & \textbf{0.212} & 0.0 & \textbf{0.111} & \textbf{0.69} $\pm$ 0.014 & 0.685 $\pm$ 0.006 \\
                         & ConvNeXt-tiny & 0.134 & \textbf{0.314} & 0.03 & \textbf{0.158} & \textbf{0.862} $\pm$ 0.007 & 0.854 $\pm$ 0.005 \\
    \midrule
    \multirow{2}{*}{ImageNet} & VGG11-bn & 0.029 & \textbf{0.217} & 0.0 & \textbf{0.01} & \textbf{0.704} & 0.699 \\
                              & ResNet-18 & 0.065 & \textbf{0.256} & 0.0 & \textbf{0.059} & \textbf{0.698} & 0.697 \\
    \bottomrule
  \end{tabular}}
  \caption{Adversarial attack results comparing original and modified models with pixel constraint and $\ell_1$ loss. The original models come from PyTorch Image Models \cite{rw2019timm} and pretrained on ImageNet.}
  \label{tab:adv_attacks_results}
\end{table}

While our approach does not achieve the robustness of adversarially trained networks, it demonstrates improved performance compared to standard networks.

\clearpage

\section{Sparsity}

For almost all datasets and architectures, our approach achieved sparser CAMs. We see especially large decreases for ImageNet.

\begin{table}[ht]
  \centering
\begin{tabular}{llll}
    \toprule
    Dataset & Architecture & $\ell_1$ $\downarrow$ & $\ell_1$ $\downarrow$ (ours) \\
    \midrule
    \multirow{3}{*}{COCO \cite{lin2015microsoft}} 
        & EfficientNet-B0 & 0.179 & \textbf{0.064} \\
        & ConvNeXt-tiny & 0.251 & \textbf{0.151} \\
        & ResNet-18 & \textbf{0.173} & 0.194 \\
    \midrule
    \multirow{2}{*}{Wood \cite{nieradzik2023automating}} 
        & EfficientNet-B0 & 0.190 & \textbf{0.032} \\
        & ConvNeXt-tiny & 0.110 & \textbf{0.046} \\
    \midrule
    \multirow{2}{*}{Oxford \cite{parkhi12a}} 
        & EfficientNet-B0 & 0.154 & \textbf{0.072} \\
        & ResNet-18 & 0.151 & \textbf{0.064} \\
    \midrule
    \multirow{3}{*}{CUB-200-2011 \cite{wah_branson_welinder_perona_belongie_2011}} 
        & EfficientNet-B0 & 0.235 & \textbf{0.05} \\
        & ConvNeXt-tiny & 0.164 & \textbf{0.096} \\
        & ResNet-18 & 0.121 & \textbf{0.056} \\
    \midrule
    \multirow{2}{*}{ImageNet \cite{5206848}} 
        & VGG11-BN & 0.279 & \textbf{0.064} \\
        & ResNet-18 & 0.387 & \textbf{0.123} \\
    \bottomrule
\end{tabular}
  \caption{The last column reports the sparsity of the CAM using our approach (with pixel constraint, $\ell_1$ loss, and changes to the model). The third column is a standard model without any changes. For the standard model, we use GradCAM.}
  \label{tab:l1}
\end{table}

Only the lowest $\ell_1$ of the different $k$ values is reported. We observe that in general a strong pixel constraint such as $k = 64$ pixels leads to the lowest $\ell_1$ value.

\clearpage

\section{Effect of $\ell_1$ normalization on robustness and interpretability}
\label{sec:detailsl1}

$\ell_1$ regularization has a strong influence on the results. Here we want to show that without the other components we would have lower interpretability, robustness and/or accuracy.

We consider the following variants of ResNet-18:

\begin{itemize}
    \item $\ell_1$ regularization only on the last feature output (activations) with $\lambda=1.0$
    \item $\ell_1$ regularization only on the last feature output (activations) with $\lambda=0.1$
    \item $\ell_1$ regularization on all activations with $\lambda=10^{-3}$
    \item $\ell_1$ regularization on all activations with $\lambda=10^{-5}$
    \item ours: our approach
\end{itemize}

We use the CUB-200-2011 dataset. $\lambda$ denotes the strength of the regularization.

\subsection{Interpretability and accuracy}

First, we analyze the effective receptive field and accuracy.

\begin{table}[ht]
  \centering
  \begin{tabular}{lllllll}
        \toprule
        Approach & $\lambda$ & Center ERF $\uparrow$ & Corner ERF $\downarrow$ & Accuracy\\\hline
        last & $1.0$ & 0.514 & \textbf{0.002} & 0.63\\
        last & $0.1$ & \textbf{0.536} & 0.113 & \textbf{0.69}\\
        all & $10^{-5}$ & 0.488 & 0.416 & \textbf{0.69}\\
        all & $10^{-3}$ & 0.335 & 0.26 & 0.19\\
        \midrule
        ours & 1.0 & \textbf{0.534} & \textbf{0.005} & \textbf{0.69}\\
        \bottomrule
        \end{tabular}
  \caption{$\ell_1$ regularization is a tradeoff between accuracy and ERF for the other approaches.}
  \label{tab:tradeoffl1}
\end{table}

We see that we are only able to influence the ERF by regularizing the last feature map. While the approach "last + $\lambda=1.0$" also achieves the same ERF as "ours", we see a significant decrease in accuracy of about 6\%. Instead, we can also decrease $\lambda$, then the accuracy is the same, but we lose interpretability.

Additionally, without our Top-GAP pooling, we can no longer control the number of pixels. The $\lambda$ parameter cannot be used for that.

\newpage

Let $X$ be the last feature output. We measure how many pixels are highlighted in the output, when adjusting $\lambda$ and our Top-GAP $k$ pixel constraint.

\begin{table}[ht]
  \centering
  \begin{tabular}{lllllll}
        \toprule
        Approach & $\lambda$ & $||X||_1$ & Accuracy $\uparrow$\\\hline
        last & $1.0$ & 0.141 & 0.63\\
        last & $0.1$ & 0.152 & \textbf{0.69}\\
        all & $10^{-5}$ & 0.119 & \textbf{0.69}\\
        all & $10^{-3}$ & 0.389 & 0.19\\
        \midrule
        ours, $k=128$ & 1.0 & 0.057 & 0.66\\
        ours, $k=256$ & 1.0 & 0.072 & 0.67\\
        ours, $k=512$ & 1.0 & 0.126 & 0.68\\
        ours, $k=1024$ & 1.0 & 0.174 & \textbf{0.69}\\
        ours, $k=2048$ & 1.0 & 0.193 & 0.68\\
        \bottomrule
        \end{tabular}
  \caption{As we increase the constraint value $k$, the number of pixels increases. The same behavior is not possible using $\lambda$. The accuracy would suffer too much.}
  \label{tab:tradeoffl12}
\end{table}

When we increase the regularization strength from $\lambda = 0.1$ to $\lambda=1.0$, the number of pixels only decreases from $0.152$ to $0.141$. However, the accuracy decreases by $6\%$.

Compare this to our approach. We can decrease the number of pixels while keeping the accuracy at the same level.

\subsection{Robustness}

Next, we analyze the level of robustness with respect to $\ell_1$ regularization.

\begin{table}[ht]
  \centering
  \begin{tabular}{lllllll}
        \toprule
        Approach & $\lambda$ & PGD$^{40}$ $\uparrow$ & FGSM\\\hline
        last & $1.0$ & 0.06 & 0.15\\
        last & $0.1$ & 0.03 & 0.11\\
        all & $10^{-5}$ & 0.0 & 0.06\\
        all & $10^{-3}$ & 0.0 & 0.03\\
        \midrule
        ours & 1.0 & \textbf{0.11} & \textbf{0.21}\\
        \bottomrule
        \end{tabular}
  \caption{Regularizing the last layer leads to the highest level of robustness. Our approach surpasses a simple regularization.}
  \label{tab:tradeoffl13}
\end{table}

Regularizing only the last layer also brings a certain degree of robustness, but it comes at a price. The accuracy is lower and we still do not achieve the same level of sparsity for $\lambda = 1.0$ as with our approach.

%% file: main.bbl
\begin{thebibliography}{10}
\providecommand{\url}[1]{\texttt{#1}}
\providecommand{\urlprefix}{URL }
\providecommand{\doi}[1]{https://doi.org/#1}

\bibitem{DBLP:conf/eccv/AndriushchenkoC20}
Andriushchenko, M., Croce, F., Flammarion, N., Hein, M.: Square attack: {A} query-efficient black-box adversarial attack via random search. In: Vedaldi, A., Bischof, H., Brox, T., Frahm, J. (eds.) Computer Vision - {ECCV} 2020 - 16th European Conference, Glasgow, UK, August 23-28, 2020, Proceedings, Part {XXIII}. Lecture Notes in Computer Science, vol. 12368, pp. 484--501. Springer (2020). \doi{10.1007/978-3-030-58592-1\_29}, \url{https://doi.org/10.1007/978-3-030-58592-1\_29}

\bibitem{andriushchenko2020understanding}
Andriushchenko, M., Flammarion, N.: Understanding and improving fast adversarial training (2020)

\bibitem{athalye2018obfuscated}
Athalye, A., Carlini, N., Wagner, D.: Obfuscated gradients give a false sense of security: Circumventing defenses to adversarial examples (2018)

\bibitem{bolukbasi2016man}
Bolukbasi, T., Chang, K.W., Zou, J., Saligrama, V., Kalai, A.: Man is to computer programmer as woman is to homemaker? debiasing word embeddings (2016)

\bibitem{pmlr-v81-buolamwini18a}
Buolamwini, J., Gebru, T.: Gender shades: Intersectional accuracy disparities in commercial gender classification. In: Friedler, S.A., Wilson, C. (eds.) Proceedings of the 1st Conference on Fairness, Accountability and Transparency. Proceedings of Machine Learning Research, vol.~81, pp. 77--91. PMLR (23--24 Feb 2018), \url{https://proceedings.mlr.press/v81/buolamwini18a.html}

\bibitem{burns2019women}
Burns, K., Hendricks, L.A., Saenko, K., Darrell, T., Rohrbach, A.: Women also snowboard: Overcoming bias in captioning models (2019)

\bibitem{byun2023reciprocam}
Byun, S.Y., Lee, W.: Recipro-cam: Fast gradient-free visual explanations for convolutional neural networks (2023)

\bibitem{9292601}
Cai, J., Hou, J., Lu, Y., Chen, H., Kneip, L., Schwertfeger, S.: Improving cnn-based planar object detection with geometric prior knowledge. In: 2020 IEEE International Symposium on Safety, Security, and Rescue Robotics (SSRR). pp. 387--393 (2020). \doi{10.1109/SSRR50563.2020.9292601}

\bibitem{Chattopadhay_2018}
Chattopadhay, A., Sarkar, A., Howlader, P., Balasubramanian, V.N.: Grad-{CAM}++: Generalized gradient-based visual explanations for deep convolutional networks. In: 2018 {IEEE} Winter Conference on Applications of Computer Vision ({WACV}). {IEEE} (mar 2018). \doi{10.1109/wacv.2018.00097}

\bibitem{clarysse2022adversarial}
Clarysse, J., Hörrmann, J., Yang, F.: Why adversarial training can hurt robust accuracy (2022)

\bibitem{5206848}
Deng, J., Dong, W., Socher, R., Li, L.J., Li, K., Fei-Fei, L.: Imagenet: A large-scale hierarchical image database. In: 2009 IEEE Conference on Computer Vision and Pattern Recognition. pp. 248--255 (2009). \doi{10.1109/CVPR.2009.5206848}

\bibitem{fang2022eva}
Fang, Y., Wang, W., Xie, B., Sun, Q., Wu, L., Wang, X., Huang, T., Wang, X., Cao, Y.: Eva: Exploring the limits of masked visual representation learning at scale (2022)

\bibitem{DBLP:journals/corr/abs-2008-02312}
Fu, R., Hu, Q., Dong, X., Guo, Y., Gao, Y., Li, B.: Axiom-based grad-cam: Towards accurate visualization and explanation of cnns. CoRR  \textbf{abs/2008.02312} (2020), \url{https://arxiv.org/abs/2008.02312}

\bibitem{https://doi.org/10.48550/arxiv.1412.6572}
Goodfellow, I.J., Shlens, J., Szegedy, C.: Explaining and harnessing adversarial examples (2014). \doi{10.48550/ARXIV.1412.6572}, \url{https://arxiv.org/abs/1412.6572}

\bibitem{goodfellow2015explaining}
Goodfellow, I.J., Shlens, J., Szegedy, C.: Explaining and harnessing adversarial examples (2015)

\bibitem{DBLP:journals/corr/abs-2110-09468}
Gowal, S., Rebuffi, S., Wiles, O., Stimberg, F., Calian, D.A., Mann, T.A.: Improving robustness using generated data. CoRR  \textbf{abs/2110.09468} (2021), \url{https://arxiv.org/abs/2110.09468}

\bibitem{he2015deep}
He, K., Zhang, X., Ren, S., Sun, J.: Deep residual learning for image recognition (2015)

\bibitem{he2023efficient}
He, Y., Yang, X., Chang, C.M., Xie, H., Igarashi, T.: Efficient human-in-the-loop system for guiding dnns attention (2023)

\bibitem{DBLP:journals/corr/abs-1903-12261}
Hendrycks, D., Dietterich, T.G.: Benchmarking neural network robustness to common corruptions and perturbations. CoRR  \textbf{abs/1903.12261} (2019), \url{http://arxiv.org/abs/1903.12261}

\bibitem{9146293}
Hou, W., Tao, X., Xu, D.: Combining prior knowledge with cnn for weak scratch inspection of optical components. IEEE Transactions on Instrumentation and Measurement  \textbf{70},  1--11 (2021). \doi{10.1109/TIM.2020.3011299}

\bibitem{huang2022exploring}
Huang, H., Wang, Y., Erfani, S.M., Gu, Q., Bailey, J., Ma, X.: Exploring architectural ingredients of adversarially robust deep neural networks (2022)

\bibitem{9462463}
Jiang, P.T., Zhang, C.B., Hou, Q., Cheng, M.M., Wei, Y.: Layercam: Exploring hierarchical class activation maps for localization. IEEE Transactions on Image Processing  \textbf{30},  5875--5888 (2021). \doi{10.1109/TIP.2021.3089943}

\bibitem{Jo_2021}
Jo, S., Yu, I.J.: Puzzle-{CAM}: Improved localization via matching partial and full features. In: 2021 {IEEE} International Conference on Image Processing ({ICIP}). {IEEE} (sep 2021). \doi{10.1109/icip42928.2021.9506058}, \url{https://doi.org/10.1109/icip42928.2021.9506058}

\bibitem{kirillov2023segment}
Kirillov, A., Mintun, E., Ravi, N., Mao, H., Rolland, C., Gustafson, L., Xiao, T., Whitehead, S., Berg, A.C., Lo, W.Y., Dollár, P., Girshick, R.: Segment anything (2023)

\bibitem{DBLP:journals/corr/abs-1802-10171}
Li, K., Wu, Z., Peng, K., Ernst, J., Fu, Y.: Tell me where to look: Guided attention inference network. CoRR  \textbf{abs/1802.10171} (2018), \url{http://arxiv.org/abs/1802.10171}

\bibitem{DBLP:journals/corr/abs-2104-14556}
Li, Z., Xu, C.: Discover the unknown biased attribute of an image classifier. CoRR  \textbf{abs/2104.14556} (2021), \url{https://arxiv.org/abs/2104.14556}

\bibitem{lin2014network}
Lin, M., Chen, Q., Yan, S.: Network in network (2014)

\bibitem{lin2017feature}
Lin, T.Y., Dollár, P., Girshick, R., He, K., Hariharan, B., Belongie, S.: Feature pyramid networks for object detection (2017)

\bibitem{lin2015microsoft}
Lin, T.Y., Maire, M., Belongie, S., Bourdev, L., Girshick, R., Hays, J., Perona, P., Ramanan, D., Zitnick, C.L., Dollár, P.: Microsoft coco: Common objects in context (2015)

\bibitem{liu2022convnet}
Liu, Z., Mao, H., Wu, C.Y., Feichtenhofer, C., Darrell, T., Xie, S.: A convnet for the 2020s (2022)

\bibitem{luo2017understanding}
Luo, W., Li, Y., Urtasun, R., Zemel, R.: Understanding the effective receptive field in deep convolutional neural networks (2017)

\bibitem{nieradzik2023automating}
Nieradzik, L., Sieburg-Rockel, J., Helmling, S., Keuper, J., Weibel, T., Olbrich, A., Stephani, H.: Automating wood species detection and classification in microscopic images of fibrous materials with deep learning (2023)

\bibitem{DBLP:journals/corr/abs-1908-01224}
Omeiza, D., Speakman, S., Cintas, C., Weldemariam, K.: Smooth grad-cam++: An enhanced inference level visualization technique for deep convolutional neural network models. CoRR  \textbf{abs/1908.01224} (2019), \url{http://arxiv.org/abs/1908.01224}

\bibitem{parkhi12a}
Parkhi, O.M., Vedaldi, A., Zisserman, A., Jawahar, C.V.: Cats and dogs. In: IEEE Conference on Computer Vision and Pattern Recognition (2012)

\bibitem{pathak2015constrained}
Pathak, D., Krähenbühl, P., Darrell, T.: Constrained convolutional neural networks for weakly supervised segmentation (2015)

\bibitem{peng2023robust}
Peng, S., Xu, W., Cornelius, C., Hull, M., Li, K., Duggal, R., Phute, M., Martin, J., Chau, D.H.: Robust principles: Architectural design principles for adversarially robust cnns (2023)

\bibitem{radford2021learning}
Radford, A., Kim, J.W., Hallacy, C., Ramesh, A., Goh, G., Agarwal, S., Sastry, G., Askell, A., Mishkin, P., Clark, J., Krueger, G., Sutskever, I.: Learning transferable visual models from natural language supervision (2021)

\bibitem{DBLP:journals/corr/abs-1906-06032}
Raghunathan, A., Xie, S.M., Yang, F., Duchi, J.C., Liang, P.: Adversarial training can hurt generalization. CoRR  \textbf{abs/1906.06032} (2019), \url{http://arxiv.org/abs/1906.06032}

\bibitem{rajabi2022fair}
Rajabi, A., Yazdani-Jahromi, M., Garibay, O.O., Sukthankar, G.: Through a fair looking-glass: mitigating bias in image datasets (2022)

\bibitem{redmon2016yolo9000}
Redmon, J., Farhadi, A.: Yolo9000: Better, faster, stronger (2016)

\bibitem{ribeiro2016why}
Ribeiro, M.T., Singh, S., Guestrin, C.: "why should i trust you?": Explaining the predictions of any classifier (2016)

\bibitem{DBLP:journals/corr/abs-1911-08731}
Sagawa, S., Koh, P.W., Hashimoto, T.B., Liang, P.: Distributionally robust neural networks for group shifts: On the importance of regularization for worst-case generalization. CoRR  \textbf{abs/1911.08731} (2019), \url{http://arxiv.org/abs/1911.08731}

\bibitem{Selvaraju_2019}
Selvaraju, R.R., Cogswell, M., Das, A., Vedantam, R., Parikh, D., Batra, D.: Grad-{CAM}: Visual explanations from deep networks via gradient-based localization. International Journal of Computer Vision  \textbf{128}(2),  336--359 (oct 2019). \doi{10.1007/s11263-019-01228-7}

\bibitem{simonyan2015deep}
Simonyan, K., Zisserman, A.: Very deep convolutional networks for large-scale image recognition (2015)

\bibitem{sun2023alternative}
Sun, W., Liu, Z., Zhang, Y., Zhong, Y., Barnes, N.: An alternative to wsss? an empirical study of the segment anything model (sam) on weakly-supervised semantic segmentation problems (2023)

\bibitem{tan2020efficientnet}
Tan, M., Le, Q.V.: Efficientnet: Rethinking model scaling for convolutional neural networks (2020)

\bibitem{wah_branson_welinder_perona_belongie_2011}
Wah, C., Branson, S., Welinder, P., Perona, P., Belongie, S.: The Caltech-UCSD Birds-200-2011 Dataset (Jul 2011)

\bibitem{7789620}
Wang, C., Siddiqi, K.: Differential geometry boosts convolutional neural networks for object detection. In: 2016 IEEE Conference on Computer Vision and Pattern Recognition Workshops (CVPRW). pp. 1006--1013 (2016). \doi{10.1109/CVPRW.2016.130}

\bibitem{wang2020scorecam}
Wang, H., Wang, Z., Du, M., Yang, F., Zhang, Z., Ding, S., Mardziel, P., Hu, X.: Score-cam: Score-weighted visual explanations for convolutional neural networks (2020)

\bibitem{DBLP:journals/corr/abs-1905-13549}
Wang, H., Ge, S., Xing, E.P., Lipton, Z.C.: Learning robust global representations by penalizing local predictive power. CoRR  \textbf{abs/1905.13549} (2019), \url{http://arxiv.org/abs/1905.13549}

\bibitem{wang2023better}
Wang, Z., Pang, T., Du, C., Lin, M., Liu, W., Yan, S.: Better diffusion models further improve adversarial training (2023)

\bibitem{rw2019timm}
Wightman, R.: Pytorch image models. \url{https://github.com/rwightman/pytorch-image-models} (2019). \doi{10.5281/zenodo.4414861}

\bibitem{DBLP:journals/corr/abs-1905-00593}
Yang, X., Wu, B., Sato, I., Igarashi, T.: Directing dnns attention for facial attribution classification using gradient-weighted class activation mapping. CoRR  \textbf{abs/1905.00593} (2019), \url{http://arxiv.org/abs/1905.00593}

\bibitem{zarandy2015overview}
Zar{\'a}ndy, {\'A}., Rekeczky, C., Szolgay, P., Chua, L.O.: Overview of cnn research: 25 years history and the current trends. In: 2015 IEEE International Symposium on Circuits and Systems (ISCAS). pp. 401--404. IEEE (2015)

\bibitem{zhao-etal-2017-men}
Zhao, J., Wang, T., Yatskar, M., Ordonez, V., Chang, K.W.: Men also like shopping: Reducing gender bias amplification using corpus-level constraints. In: Proceedings of the 2017 Conference on Empirical Methods in Natural Language Processing. pp. 2979--2989. Association for Computational Linguistics, Copenhagen, Denmark (Sep 2017). \doi{10.18653/v1/D17-1323}, \url{https://aclanthology.org/D17-1323}

\bibitem{zhou2015learning}
Zhou, B., Khosla, A., Lapedriza, A., Oliva, A., Torralba, A.: Learning deep features for discriminative localization (2015)

\bibitem{8316924}
Zhou, X., Zhu, M., Pavlakos, G., Leonardos, S., Derpanis, K.G., Daniilidis, K.: Monocap: Monocular human motion capture using a cnn coupled with a geometric prior. IEEE Transactions on Pattern Analysis and Machine Intelligence  \textbf{41}(04),  901--914 (apr 2019). \doi{10.1109/TPAMI.2018.2816031}

\end{thebibliography}
